\documentclass[10pt,twocolumn,letterpaper]{article}

\usepackage{iccv}
\usepackage{times}
\usepackage{epsfig}
\usepackage{graphicx}
\usepackage{amsmath}
\usepackage{amssymb}
\usepackage{multicol}
\usepackage{multirow}
\usepackage{authblk}

\usepackage[accsupp]{axessibility}  

\usepackage[pagebackref=true,breaklinks=true,letterpaper=true,colorlinks,bookmarks=false]{hyperref}

\usepackage[capitalize]{cleveref}
\crefname{section}{Sec.}{Secs.}
\Crefname{section}{Section}{Sections}
\Crefname{table}{Table}{Tables}
\crefname{table}{Tab.}{Tabs.}

\iccvfinalcopy 

\newcommand{\xu}{\color{black}}
\newcommand{\cheng}{\color{black}}
\newcommand{\final}{\color{black}}

\def\iccvPaperID{1356} 

\ificcvfinal\fi

\begin{document}

\title{Self-Calibrated Cross Attention Network for Few-Shot Segmentation}

\author{
Qianxiong Xu\textsuperscript{1}\quad
Wenting Zhao\textsuperscript{2}\quad
Guosheng Lin\textsuperscript{1}\footnote[1]{}\quad
Cheng Long\textsuperscript{1}\thanks{Co-corresponding authors}\\
\textsuperscript{1}S-Lab, Nanyang Technological University\quad
\textsuperscript{2}Nanjing University of Science and Technology\\
{\tt\small \{qianxiong.xu,gslin,c.long\}@ntu.edu.sg, wtingzhao@njust.edu.cn}
}

\maketitle
\ificcvfinal\thispagestyle{empty}\fi

\begin{abstract}
    The key to the success of few-shot segmentation (FSS) lies in how to effectively utilize support samples. Most solutions compress support foreground (FG) features into prototypes, but lose some spatial details. Instead, others use cross attention to fuse query features with uncompressed support FG. Query FG could be fused with support FG, however, query background (BG) cannot find matched BG features in support FG, yet inevitably integrates dissimilar features. Besides, as both query FG and BG are combined with support FG, they get entangled, thereby leading to ineffective segmentation. To cope with these issues, we design a self-calibrated cross attention (SCCA) block. For efficient patch-based attention, query and support features are firstly split into patches. Then, we design a patch alignment module to align each query patch with its most similar support patch for better cross attention. Specifically, SCCA takes a query patch as $Q$, and groups the patches from the same query image and the aligned patches from the support image as $K\&V$. In this way, the query BG features are fused with matched BG features (from query patches), and thus the aforementioned issues will be mitigated. Moreover, when calculating SCCA, we design a scaled-cosine mechanism to better utilize the support features for similarity calculation. Extensive experiments conducted on PASCAL-5$^i$ and COCO-20$^i$ demonstrate the superiority of our model, \eg, the mIoU score under 5-shot setting on COCO-20$^i$ is 5.6\%+ better than previous state-of-the-arts. The code is available at \url{https://github.com/Sam1224/SCCAN}.
\end{abstract}

\section{Introduction}
\label{sec:intro}

%
%
%
%
%
%

With the rapid development of deep learning,
semantic segmentation has made tremendous progress~\cite{long2015fully,zhao2017pyramid,chen2017deeplab},
but such success requires massive time and human efforts for annotating the pixel-wise masks.
To save annotation cost, semi-/weakly-supervised segmentation~\cite{wei2018revisiting,jiang2022l2g,fan2022ucc} are proposed to take advantages of unlabelled/weakly-labelled data. 
Nevertheless, they both fail to segment unseen classes (during training), which prevents segmentation from generalizing to novel classes.
To tackle this problem, few-shot segmentation (FSS)~\cite{shaban2017one,wang2019panet,zhang2019canet} is introduced to quickly adapt the model to novel classes, with the help of a few annotated samples.

\begin{figure}[t]
  \begin{center}
  \includegraphics[width=1\linewidth]{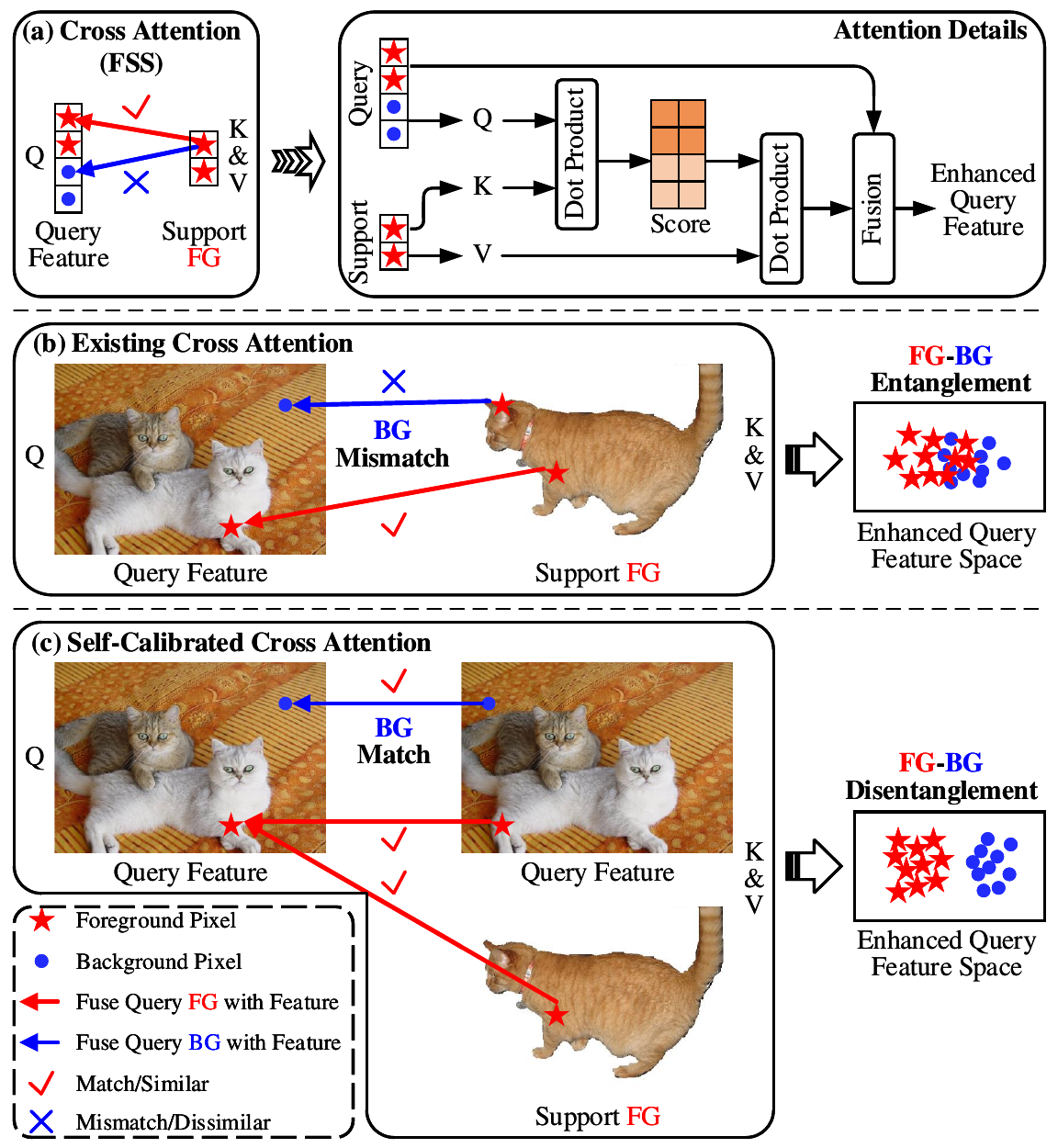}
  \end{center}
  \caption{{\bf (a) Details of cross attention for FSS. (b)\&(c) Existing and our proposed cross attentions.}
  In {\bf (b)}, query {\color{red} FG} features are correctly fused with matched support {\color{red} FG} features, but query {\color{blue} BG} features are combined with mismatched support {\color{red} FG} features, and they get entangled.
  In {\bf (c)}, query {\color{blue} BG} is correctly fused with matched query {\color{blue} BG} instead, so query features are disentangled.}
  \label{fig:framework_comparisons}
  \vspace{-1em}
\end{figure}

Naturally, human could refer to a few samples of a novel object, and recognize them in unlabelled images, even without knowing what they actually are, \ie, class-agnostic.
Inspired by this phenomenon, researchers develop the learning-to-learn paradigm~\cite{ren2018meta,snell2017prototypical} for FSS models. They learn a segmentation pattern on some base classes during training, and apply the same pattern to segment novel classes during testing.
Particularly, this pattern is to use the knowledge of a few annotated support samples to perform segmentation on an unlabelled query image for a class.

There exist many solutions for FSS, where prototype-based methods~\cite{zhang2019canet,wang2019panet,li2021adaptive} are the most popular.
Specifically, support prototypes~\cite{zhang2019canet,li2021adaptive} are extracted from support FG\footnote{{\bf FG} and {\bf BG} are used to represent foreground and background.}, and they are used to segment query image through, \eg, feature comparison~\cite{wang2019panet}.
However, compressing support FG to prototypes leads to information loss, 
Therefore, some methods~\cite{zhang2019pyramid,wang2020few,zhang2021few,wang2022adaptive} employ cross attention instead to fuse query features with uncompressed support FG (as shown in \cref{fig:framework_comparisons}(a)).
Unfortunately, two other issues would arise.
(1) As shown in \cref{fig:framework_comparisons}(b), query BG ($Q$, carpet) cannot find matched features in support FG ($K\&V$, cat), \ie, the similarity scores between all query BG and support FG pixel pairs are low. Then, such low scores are amplified after softmax normalization, \eg, 1e-10 to 0.1 for one pair.
As a result, query BG features are inevitably fused with mismatched support FG features and get biased.
This issue is termed {\em BG mismatch}.
(2) Meanwhile, query FG could effectively integrate the knowledge of matched support FG. As both query FG and BG are fused with support FG, their features get entangled,  which is called {\em FG-BG entanglement}, thereby leading to ineffective segmentation.

To address these two issues, we propose a novel attention module named {\bf self-calibrated cross attention (SCCA)} that calculates self and cross attentions simultaneously.
As shown in \cref{fig:framework_comparisons}(c), query features are still taken as $Q$, but the query features and support FG are grouped as $K\&V$. The rationality is explained as follows:
(1) Query BG features are effectively fused with matched query BG features from the same query image in $K\&V$ (via self attention). Hence, the {\em BG mismatch} issue is solved.
(2) Query FG can be enhanced by matched information in both query features (via self attention) and support FG (via cross attention). In this way, query BG features persist to integrate BG information, while query FG is combined with FG information, the {\em FG-BG entanglement} issue is mitigated.

Unluckily, the attention cost of SCCA is twice as much as that of standard cross attention.
Therefore, we incorporate SCCA with the memory efficient swin transformer (ST)~\cite{liu2021swin}, and present the {\bf self-calibrated cross attention network (SCCAN)} to boost FSS.
To be specific, a pseudo mask aggregation (PMA) module is firstly developed to generate pseudo query masks. PMA takes \emph{all} pairwise similarities to generate the mask value for a query pixel, which is better at alleviating the effects of noises, compared to existing methods~\cite{luo2021pfenet++,tian2020prior} that merely use a \emph{single} largest similarity value.
Then, {\bf self-calibrated cross attention (SCCA) block} is proposed for effective patch-based cross attention, which mainly consists of (1) patch alignment (PA) module and (2) aforementioned SCCA module.
(1) ST-based cross attention may have the patch misalignment issue in local attention calculation.
Thus, we design a PA module to align each query patch with its most similar support patch (with FG pixels).
(2) Given pairs of query and aligned support patches, SCCA effectively fuses query features with support FG features.
Recall that query FG features integrate information both from the same query image patches and the support image patches. Naturally, a query patch is more similar to itself than another support patch, and thus self attention might 
dominate
in SCCA, and the query FG features may not integrate sufficient support FG features, which may weaken the FG feature representations. Hence, we further incorporate SCCA with a scaled-cosine (SC) mechanism to encourage query FG to integrate more information from the support image.
In a nutshell, our contributions could be summarized as follows:
\begin{itemize}
    \item We propose a self-calibrated cross attention network (SCCAN), including pseudo mask aggregation (PMA) module and self-calibrated cross attention (SCCA) blocks, to effectively utilize support information.
    \item SCCA could tackle the {\em BG mismatch} and {\em FG-BG entanglement} issues, which disentangles query FG and BG, thereby leading to effective segmentation.
    \item Training-agnostic PMA module could roughly locate query FG, and it is better at suppressing the effects of noises, compared to existing methods.
    \item New state-of-the-art could be set on PASCAL-5$^i$ and COCO-20$^i$, \eg, the mIoU score under 5-shot setting on COCO-20$^i$ is 5.6\%+ better than previous methods.
\end{itemize}

\section{Related work}
\label{sec:related_work}

%
%
%

\begin{figure*}[t]
  \begin{center}
  \includegraphics[width=1\linewidth]{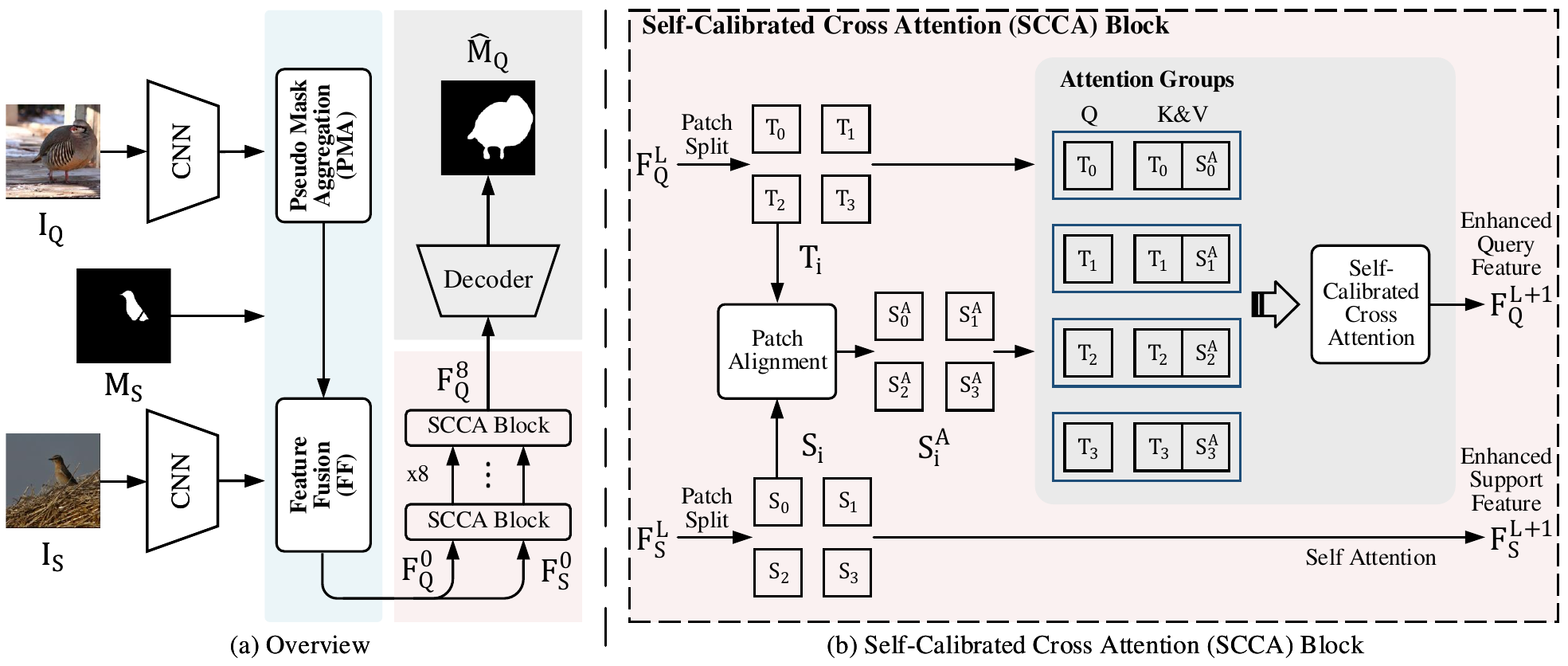}
  \end{center}
  \caption{{\bf Overall architecture of (a) self-calibrated cross attention network (SCCAN) and (b) self-calibrated cross attention (SCCA) block.}
  {\bf (a) Pseudo mask aggregation} module generates a pseudo mask that could roughly locate query FG.
  {\bf Feature fusion} module adapts query and support features for better cross attention.
  {\bf SCCA blocks} could effectively fuse query features with support FG.
  Finally, the enhanced query features is forwarded to a decoder for segmentation.
  {\bf (b) SCCA blocks} perform patch-based attentions. {\bf Patch alignment} module aims at finding the most similar support patch $S_i^A$ for each query patch $T_i$. {\bf SCCA} takes a query patch $T_i$ as $Q$, and groups the same $T_i$ and its aligned support patch $S_i^A$ as $K\&V$ to address the {\em BG mismatch} and {\em FG-BG entanglement} issues.
  }
  \label{fig:overview}
\end{figure*}

\smallskip
\noindent
\textbf{Few-shot segmentation.}
To alleviate the 
generalization
problem on unseen classes of semantic segmentation, few-shot segmentation (FSS) is firstly introduced in OSLSM~\cite{shaban2017one}, 
which segments
query image with the help of annotated support samples.
Recent FSS methods could be roughly divided into four categories.
(1) Prototype-based methods~\cite{zhang2019canet,wang2019panet,tian2020prior,li2021adaptive,lang2022beyond,liu2022learning,zhang2021self,fan2022self,liu2022intermediate} compress support FG into single or multiple prototypes, which are then used to help segment the query image through, \eg, cosine similarity or feature concatenation.
{\xu
{\cheng Based on the intuition}
that pixels from the same object are more similar than those from {\cheng different} objects, SSP~\cite{fan2022self} {\cheng is designed to generate} query FG and BG prototypes to find other similar query features.
}
(2) To prevent from information loss of prototypes, attention-based methods~\cite{zhang2021few,zhang2019pyramid,wang2020few,hu2019attention,xie2021scale,wang2022adaptive,jiao2022mask} build up per-pixel relationships between query and uncompressed support features, then use cross attention to fuse query features with support FG features. Nevertheless, 
they suffer from the {\em BG mismatch} and {\em FG-BG entanglement} 
issues as explained in \cref{sec:intro}.
(3) More recently, correlation-based methods~\cite{min2021hypercorrelation, shi2022dense, hong2022cost,xiong2022doubly} focus on visual correspondence to build 4D correlations upon query and support features, then use expensive 4D operations to perform segmentation.
(4) There also exist some other methods, \eg, DPCN~\cite{liu2022dynamic} generates dynamic kernels from support sample which is then used to process query features, some methods~\cite{lang2022learning,sun2022singular,iqbal2022msanet} benefit from base classes during testing and generate good results.
In this work, we focus on the cross attention-based methods,
and contribute effective self-calibrated cross attention (SCCA) for FSS, to solve the aforementioned issues.

\smallskip
\noindent
\textbf{Attention.}
Recently, some works~\cite{dosovitskiy2020image,touvron2021training,wang2021pyramid,yuan2021tokens,zhang2022hivit,liu2021swin} demonstrate that pure transformer architecture could achieve excellent results in computer vision tasks.
Particularly, swin transformer (ST)~\cite{liu2021swin} calculates efficient self attention within small windows to reduce computational burden, while achieving good results.
In spite of its good performance, ST in its original form does not support effective cross attention that could be utilized to fuse query features with support samples in FSS. 
In this work,
we adapt ST for SCCA to enable efficient and effective cross attentions.

\section{Problem definition}
\label{sec:problem_definition}

Suppose the sets for training and testing are denoted as $\mathcal{D}_{tr}$ and $\mathcal{D}_{te}$, respectively.
$\mathcal{D}_{tr}$ involves some classes $\mathcal{C}_{tr}$, while $\mathcal{D}_{te}$ covers another class set $\mathcal{C}_{te}$. FSS studies a scenario where $\mathcal{C}_{tr}$ and $\mathcal{C}_{te}$ are disjoint, \ie, $\mathcal{C}_{tr} \cap \mathcal{C}_{te} = \emptyset$.
Both $\mathcal{D}_{tr}$ and $\mathcal{D}_{te}$ contain numerous {\em episodes}, which are the basic elements of episodic training.
For $k$-shot setting, each episode consists of a support set $\mathcal{S} = \{I_S^n,M_S^n\}_{n=1}^{k}$ and a query set $\mathcal{Q} = \{I_Q,M_Q\}$ for a specific class $c$, where $I_S^n$ and $M_S^n$ represent the $n$-th support image and its annotated binary mask, $I_Q$ and $M_Q$ indicate the query image and corresponding mask.
During training, the model learns to segment $I_Q$ under the guidance of support set $\mathcal{S}$ for classes $\mathcal{C}_{tr}$, and then apply the learned pattern to $\mathcal{C}_{te}$ during testing.
For simplicity, we introduce our method under $1$-shot setting.

\section{Methodology}
\label{sec:methodology}

%
%
%
%

As shown in \cref{fig:overview}(a), we propose self-calibrated cross attention network (SCCAN), which consists of pseudo mask aggregation (PMA) module, feature fusion (FF) module and self-calibrated cross attention (SCCA) blocks.
PMA is responsible for generating a pseudo query mask that could roughly locate query FG with minor cost.
Then, FF adapts query and support features to close the gap between their FG features for better cross attention.
Particularly, SCCA is designed to alleviate the {\em BG mismatch} and {\em FG-BG entanglement} issues of existing cross attention-based FSS methods, and to effectively fuse query features with support FG information.
Next, we will introduce the details of SCCA first, followed by PMA and FF modules.

\subsection{Self-calibrated cross attention block}
\label{sec:self_calibrated_cross_attention}

\begin{figure}[t]
  \begin{center}
  \includegraphics[width=1\linewidth]{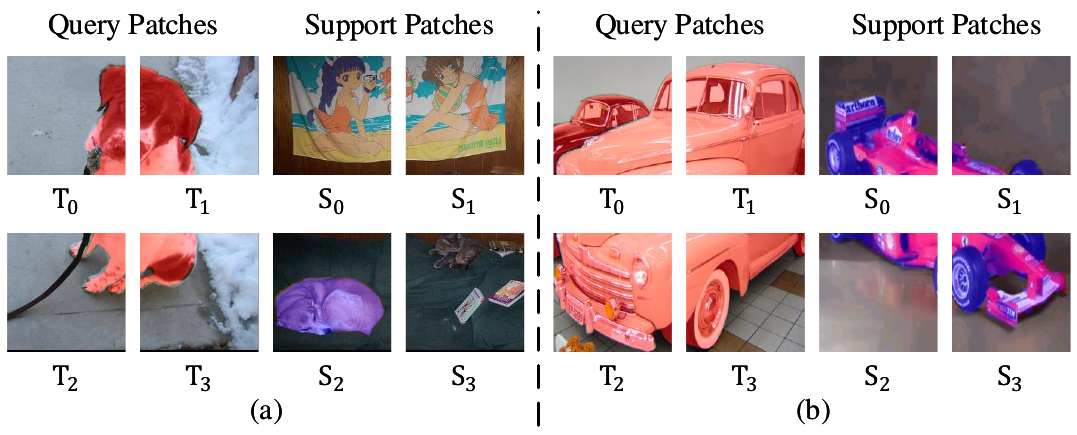}
  \end{center}
  \caption{{\bf Issues of swin transformer when used for cross attention.}
  Support BGs are usually not used in FSS, we preserve them for completeness.
  {\bf (a)} {\em Invalid support patch}.
  {\bf (b)} {\em Misaligned support patch}.
  }
  \label{fig:issue_swin}
  \vspace{-1em}
\end{figure}

Swin transformer (ST)~\cite{liu2021swin} is well known for its excellent performance and efficiency for self attention. Specifically, it seamlessly splits features into small windows, and performs window-based self attention. Then, windows are shifted for another self attention to capture long-range dependencies.
Nevertheless, merely taking query patch (we denote that pixels in a window form a patch) as $Q$, and its corresponding support patch as $K\&V$ for cross attention would raise some problems.
(1) 
{\em BG mismatch} and {\em FG-BG entanglement} (as explained earlier);
(2) {\em Invalid support patch} (\cref{fig:issue_swin}(a)): $T_0$'s corresponding patch $S_0$ does not contain FG objects, and thus $T_0$ cannot be enhanced after performing cross attention with $S_0$;
(3) {\em Misaligned support patch} (\cref{fig:issue_swin}(b)): $T_2$ represents headstock, but in support image, headstock locates in $S_3$. $T_2$ would perform cross attention with $S_2$, {and thus} its most similar features, \ie, headstock, cannot be directly accessed, which degrades the effectiveness of cross attention.

Therefore, we propose self-calibrated cross attention (SCCA) to adapt ST for effective cross attention and mitigate these issues.
We employ 8 consecutive SCCA blocks (as shown in \cref{fig:overview}(b)), each of which takes the enhanced query and support features from previous block as inputs, and seamlessly split them into patches.
Then, patch alignment (PA) module is proposed to align each query patch with its most similar support patch.
After that, each query patch and its aligned support patch are grouped for SCCA to effectively enhance query features with support FG.

The inputs for the first block come from feature fusion (FF) module, which will be introduced later.
Both input features of the $L$-th block $F_Q^{L} \in \mathbb{R}^{C \times H \times W}$ and $F_S^{L} \in \mathbb{R}^{C \times H \times W}$ are split into patches $T_i \in \mathbb{R}^{N^2 \times C \times K \times K}$ and $S_i \in \mathbb{R}^{N^2 \times C \times K \times K}, i \in \{0,\cdots,N^2-1\}$, where $N^2$ is the number of patches, and $K \times K$ denotes window/patch size.

\begin{figure}[t]
  \begin{center}
  \includegraphics[width=1\linewidth]{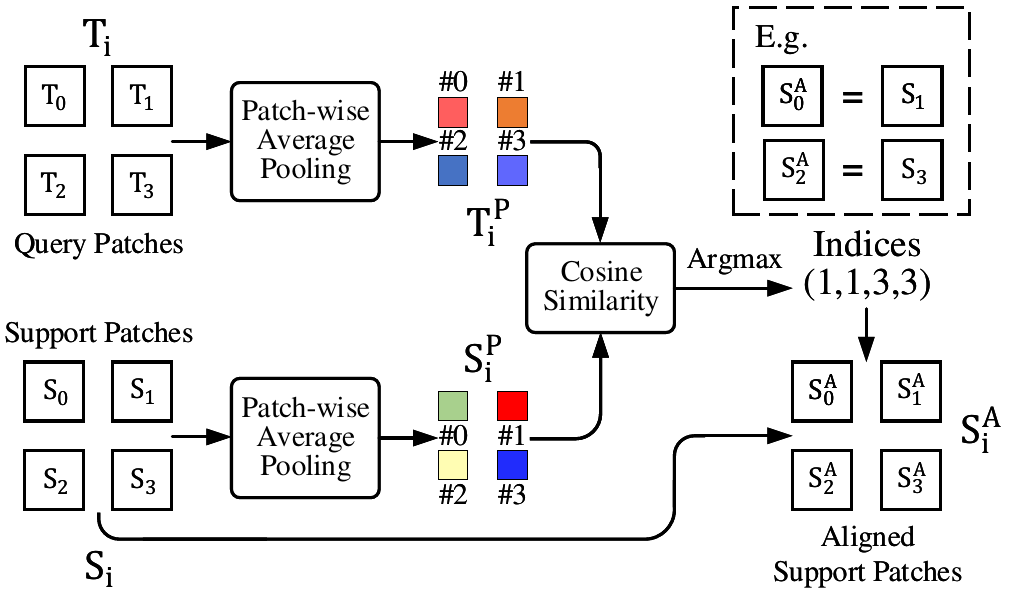}
  \end{center}
  \caption{{\bf Details of patch alignment (PA) module.}
  Firstly, prototype is obtained from each input patch.
  Then, cosine similarity is calculated among query and support prototypes.
  Finally, each query patch's most similar support patch is obtained via $\arg\max$ operation, and then they are aligned.
  }
  \label{fig:patch_alignment}
  \vspace{-1em}
\end{figure}

\smallskip
\noindent
\textbf{Patch alignment (PA).}
PA could mitigate the {\em invalid support patch} and {\em misaligned support patch} issues by aligning each query patch with its most similar support patch (with FG pixels) for effective cross attention.
Specifically, as shown in \cref{fig:patch_alignment}, the prototype of each patch is obtained via patch-wise average pooling.
\begin{equation}
    T_i^P = PAP(T_i), S_i^P = PAP(S_i)
\end{equation}
where $T_i^P \in \mathbb{R}^{N^2 \times C \times 1 \times 1}$ and $S_i^P \in \mathbb{R}^{N^2 \times C \times 1 \times 1}$ are the prototypes of query and support patches, and $PAP$ denotes patch-wise average pooling. Note that support features and support prototypes only contain FG information.

Next, we measure the cosine similarity between patch prototypes $T_i^P$ and $S_i^P$, mask out support patches without FG objects
in
the similarity score, and perform $\mathop{\arg\max}$ to obtain each query patch's most similar support patch.
\begin{equation}
    Indices = \arg\max(Cos(T_i^P, S_i^P) \circ M_S^P)
\end{equation}
where $Indices \in \mathbb{R}^{N^2}$ are the indices of aligned support patches $S_i^{A}$, $Cos$ means cosine similarity, $\circ$ denotes Hadamard product, $M_S^P \in \{1, 0 \}^{1 \times N^2}$ is a mask indicating if each support patch has FG pixels or not.
Then, each query and its aligned support patch are grouped for SCCA.

\smallskip
\noindent
\textbf{Self-calibrated cross attention (SCCA).}
Given a pair of query and its aligned support patch, the query patch is taken as $Q$, while both of them are taken as $K\&V$, and thus self attention (query patch in $Q$ with query patch in $K\&V$) and cross attention (query patch in $Q$ with support patch in $K\&V$) would be calculated simultaneously.
We explain the rationale as follows.

\begin{figure}[t]
  \begin{center}
  \includegraphics[width=1\linewidth]{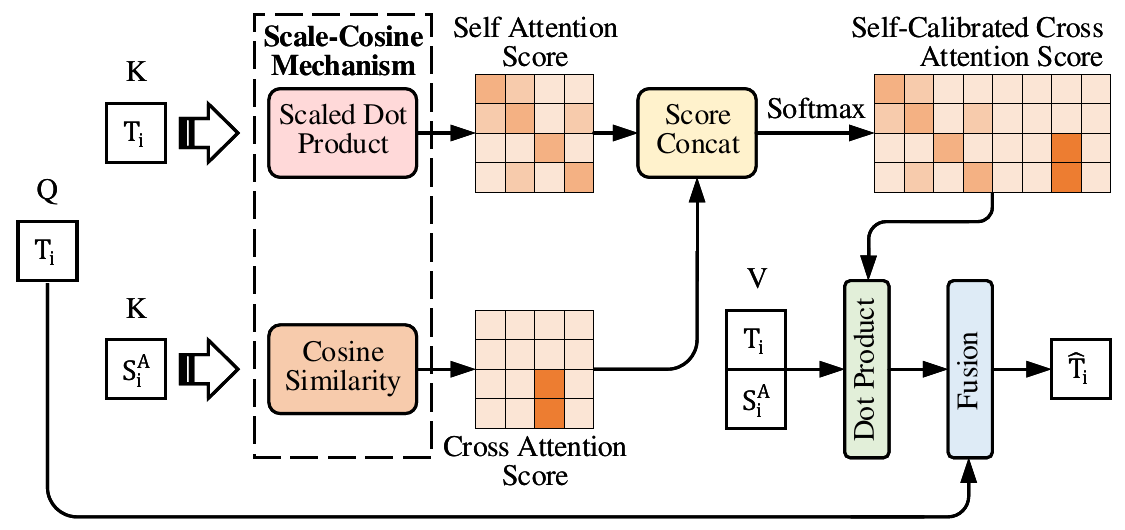}
  \end{center}
  \caption{{\bf Details of self-calibrated cross attention.}
  Fusion module is a feed-forward network.
  Query patch firstly calculates self and cross attention scores with itself and the aligned support patch.
  Scores are then concatenated and normalized.
  Finally, the score is used to obtain useful support information, which is fused with input query patch for query feature enhancement.
  }
  \label{fig:self_calibrated_cross_attention}
  \vspace{-1em}
\end{figure}

(1) For a query BG pixel, it cannot find similar information from support FG in existing cross attention-based FSS methods. 
During SCCA,
query BG could find other matched BG information from query features themselves (via self attention), while ignoring the dissimilar support FG. Thus, the {\em BG mismatch} problem is solved.

(2) Query FG could find matched FG features not only from support FG but also from query features, and thus it could be fused with reasonable features, and get enhanced.

(3) Since both query FG and BG are correctly fused with their matched features in SCCA, the {\em FG-BG entanglement} problem could be mitigated.

Recall that the performance of FSS is affected by how the support sample is utilized. Unfortunately, in (2), query FG may not be fused with sufficient support features, because FG pixels in a query patch are more similar to themselves, compared with those in the aligned support patch. As a result, self attention might dominate in SCCA.
To address this issue,
we further contribute a {\bf scaled-cosine (SC)} mechanism.
As illustrated in \cref{fig:self_calibrated_cross_attention}, when performing SCCA, we use scaled dot product for self attention, and cosine similarity for cross attention.
Their attention scores are then concatenated and normalized.
Dot product is a more strict
similarity-related operator,
as it takes both direction and magnitude into consideration.
In contrast,
cosine similarity only measures the similarity of directions.
Hence, we employ cosine similarity for cross attention to encourage
{\xu to discover more similar regions,}
and query FG is more likely to integrate sufficient information from support FG.

We formally describe SCCA as follows.
Scaled dot product is firstly conducted on query patch $T_i$ and itself to obtain self attention score.
Meanwhile, cosine similarity is performed on the same $T_i$ and aligned support patch $S_i^{A}$ to obtain cross attention score.
Their scores are denoted as $A_{QQ} \in \mathbb{R}^{N^2 \times K^2 \times K^2}$ and $A_{QS} \in \mathbb{R}^{N^2 \times K^2 \times K^2}$, where $N^2$ and $K^2$ are the number of patches and pixels, respectively.
\begin{equation}
    A_{QQ} = \frac{T_i \cdot T_i}{\sqrt{d_k}}, A_{QS} = \frac{T_i \cdot S_i^A}{\Vert T_i \Vert\Vert S_i^A \Vert}
\end{equation}
Then, two scores are concatenated and normalized by the softmax operation to obtain the final attention score $A \in \mathbb{R}^{N^2 \times K^2 \times 2K^2}$.
\begin{equation}
    A = Softmax(Concat(A_{QQ}, A_{QS}))
\end{equation}
Finally, $A$ is used to aggregate information from $T_i$ and $S_i^A$, and then be fused with $T_i$ for query feature enhancement.
\begin{equation}\label{eq:fuse}
    \hat{T}_i = FFN(A \cdot Concat(T_i, S_i^A) + T_i)
\end{equation}
where $FFN$ is feed-forward network.

Once all enhanced query patches are obtained, they are assembled to be $F_Q^{L+1} \in \mathbb{R}^{C \times H \times W}$.
Moreover, $F_Q^7$ from the last SCCA block is forwarded to the decoder~\cite{lang2022learning} and generate the segmentation $\hat{M}_Q \in \mathbb{R}^{2 \times H \times W}$.

\smallskip
\noindent
\textbf{Window shifting.}
To enable interactions among patches, we adopt window shifting operation in even layers as swin transformer~\cite{liu2021swin}.
After window shifting, some irregular patches are obtained at image borders, and we pad them to have regular size for PA and SCCA.

\subsection{Pseudo mask aggregation}
\label{sec:pseudo_mask_aggregation}

Pseudo masks~\cite{liu2022dynamic,luo2021pfenet++,tian2020prior,zhang2021few} are commonly incorporated into FSS models because they could roughly locate the query FG object without learnable parameters.
Specifically, they measure the cosine similarity between each pair of high-level query and support FG pixels~\cite{tian2020prior,zhang2021few} or patches~\cite{liu2022dynamic,luo2021pfenet++}.
Then, query pixel's largest similarity score is normalized and taken as its probability of being FG.

However, existing methods suffer from the following two issues.
(1) They only take support FG into consideration, which may not work well when both query FG and BG look dissimilar to support FG, \ie, their scores are similar.
(2) As each value in pseudo mask merely corresponds to the largest cosine similarity among pixel or patch pairs, they would be heavily affected by noises.
As a result, the locating function of pseudo mask is weakened.

\begin{figure}[t]
  \begin{center}
  \includegraphics[width=1\linewidth]{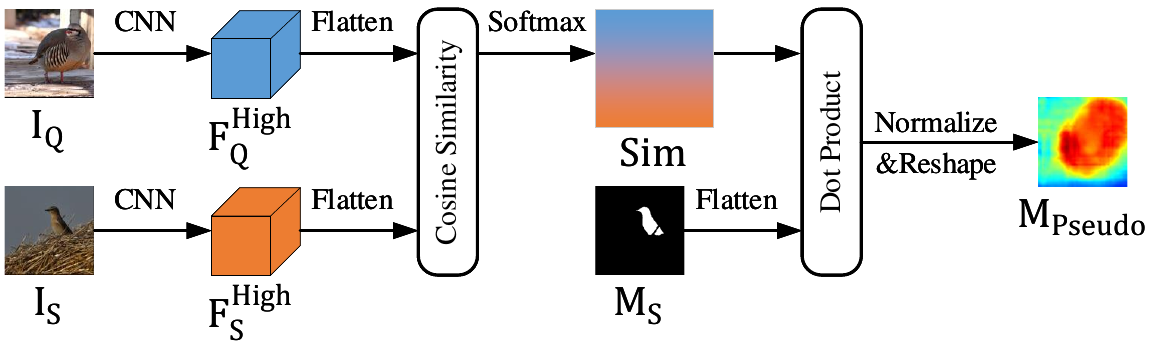}
  \end{center}
  \caption{{\bf Details of pseudo mask aggregation (PMA).}
  Given a query pixel, the weighted summation of its normalized similarity scores with all support pixels and the annotated support mask means the probability of being FG.
  }
  \label{fig:pseudo_mask_aggregation}
  \vspace{-1em}
\end{figure}

To tackle these issues, we propose a pseudo mask aggregation (PMA) module, which is illustrated in \cref{fig:pseudo_mask_aggregation}. Concretely, we apply two changes to the original one:
(1) Both support FG and BG features are used for reliable comparisons, \eg,
human head (query FG) is more similar to body (support FG) than room (support BG), and thus it is classified as FG.
(2) To suppress the effects of noises, we use all pairwise similarities to perform weighted summation on binary support mask, instead of merely taking a largest value.

First of all, PMA calculates cosine similarity $Sim \in \mathbb{R}^{HW \times HW}$ between high-level query features $F_Q^{High} \in \mathbb{R}^{C \times H \times W}$ and support features $F_S^{High} \in \mathbb{R}^{C \times H \times W}$.
Then, a softmax operation is applied to normalize the score.
After that, dot product is conducted on $Sim$ and support mask $M_S$ to obtain the pseudo mask $M_{Pseudo} \in \mathbb{R}^{HW \times 1}$.
Finally, normalization and reshape operations are employed to generate the pseudo mask $M_{Pseudo} \in \mathbb{R}^{1 \times H \times W}$.

\subsection{Feature fusion}
\label{sec:feature_fusion}

Due to the fact that FG objects in query and support images could be dissimilar, we start from the mid-level query features $F_Q^{Mid}$ and support features $F_S^{Mid}$ {\final (obtained from the $3^{rd}$ and $4^{th}$ blocks of the backbone~\cite{tian2020prior})}, and use the support prototype $P_S$ (extracted from support FG) to close their gaps.
Concretely, $F_Q^{Mid}$ is combined with $P_S$ and its pseudo mask $M_{Pseudo}$ to obtain $F_Q$.
Similarly, $F_S^{Mid}$ is fused with support prototype $P_S$ and support mask $M_S$ to generate $F_S$.
Particularly, $F_Q$ and $F_S$ are inputted to SCCA as $F_Q^0$ and $F_S^0$, respectively.

\section{Experiments}
\label{sec:experiments}

%
%
%

\subsection{Experimental settings}
\label{sec:experimental_settings}

\begin{table*}[t]
  \begin{center}
  \resizebox{.9\linewidth}{!}{
      \begin{tabular}{c|r|cccccc|cccccc}
        \hline
        \multirow{2}{*}{Backbone} & \multirow{2}{*}{Method} & \multicolumn{6}{c|}{1-shot} & \multicolumn{6}{c}{5-shot}\\
        & & 5$^0$ & 5$^1$ & 5$^2$ & 5$^3$ & Mean & FB-IoU & 5$^0$ & 5$^1$ & 5$^2$ & 5$^3$ & Mean & FB-IoU \\
        \hline
        \multirow{10}{*}{ResNet50} & PFENet$^{\dagger}$ (TPAMI'20)~\cite{tian2020prior} & 61.7 & 69.5 & 55.4 & 56.3 & 60.8 & 73.3 & 63.1 & 70.7 & 55.8 & 57.9 & 61.9 & 73.9\\
         & MLC$^{\dagger}$ (ICCV'21)~\cite{yang2021mining} & 59.2 & 71.2 & 65.6 & 52.5 & 62.1 & - & 63.5 & 71.6 & \bf 71.2 & 58.1 & 66.1 & -\\
         & HSNet$^{\ddagger\star}$ (ICCV'21)~\cite{min2021hypercorrelation} & 63.5 & 68.2 & 62.4 & \underline{59.8} & 63.5 & 76.5 & 70.1 & 72.0 & 67.9 & \bf 67.1 & 69.3 & 80.6\\
         & CyCTR$^{\dagger}$ (NIPS'21)~\cite{zhang2021few} & 65.7 & 71.0 & 59.5 & 59.7 & 64.0 & - & 69.3 & 73.5 & 63.8 & 63.5 & 67.5 & -\\
         & NTRENet$^{\dagger}$ (CVPR'22)~\cite{liu2022learning} & 65.4 & 72.3 & 59.4 & \underline{59.8} & 64.2 & 77.0 & 66.2 & 72.8 & 61.7 & 62.2 & 65.7 & 78.4\\
         & DPCN$^{\dagger}$ (CVPR'22)~\cite{liu2022dynamic} & - & - & - & - & 65.7 & \underline{77.4} & - & - & - & - & - & -\\
         & DCAMA$^{\ddagger\star}$ (ECCV'22)~\cite{shi2022dense} & 66.1 & 71.9 & 59.7 & 57.5 & 63.8 & 75.7 & 70.7 & 72.9 & 63.0 & 65.0 & 67.9 & 79.4\\
         & VAT$^{\ddagger\star}$ (ECCV'22)~\cite{hong2022cost} & 65.9 & 70.0 & 64.9 & 59.0 & 65.0 & \underline{77.4} & \underline{72.1} & \underline{74.2} & 69.7 & 65.3 & \underline{70.3} & \underline{81.1}\\
        \cline{2-14}
         & SCCAN$^{\dagger}$ (Ours) & \underline{67.5} & \bf 72.6 & \bf 67.2 & \bf 60.5 & \bf 67.0 & 76.4 & 69.9 & \bf 74.3 & \underline{70.1} & \underline{66.9} & \underline{70.3} & 79.7\\
         & SCCAN$^{\ddagger}$ (Ours) & \bf 68.3 & \underline{72.5} & \underline{66.8} & \underline{59.8} & \underline{66.8} & \bf 77.7 & \bf 72.3 & 74.1 & 69.1 & 65.6 & \bf 70.3 & \bf 81.8\\
        \hline
        \multirow{9}{*}{ResNet101} & PFENet$^{\dagger}$ (TPAMI'20)~\cite{tian2020prior} & 60.5 & 69.4 & 54.4 & 55.9 & 60.1 & 72.9 & 62.8 & 70.4 & 54.9 & 57.6 & 61.4 & 73.5\\
         & MLC$^{\dagger}$ (ICCV'21)~\cite{yang2021mining} & 60.8 & 71.3 & 61.5 & 56.9 & 62.6 & - & 65.8 & 74.9 & \bf 71.4 & 63.1 & 68.8 & -\\
         & HSNet$^{\ddagger\star}$ (ICCV'21)~\cite{min2021hypercorrelation} & 65.7 & 70.3 & 63.2 & \underline{61.9} & 65.3 & 77.2 & 72.0 & 73.6 & 68.7 & \underline{68.4} & 70.7 & 80.9\\
         & CyCTR$^{\dagger}$ (NIPS'21)~\cite{zhang2021few} & 67.2 & 71.1 & 57.6 & 59.0 & 63.7 & - & 71.0 & 75.0 & 58.5 & 65.0 & 67.4 & -\\
         & NTRENet$^{\dagger}$ (CVPR'22)~\cite{liu2022learning} & 65.5 & 71.8 & 59.1 & 58.3 & 63.7 & 75.3 & 67.9 & 73.2 & 60.1 & 66.8 & 67.0 & 78.2\\
         & DCAMA$^{\ddagger}$ (ECCV'22)~\cite{shi2022dense} & 62.5 & 70.8 & 64.5 & 56.4 & 63.5 & 76.5 & 70.0 & 73.8 & 66.8 & 65.0 & 68.9 & 81.1\\
         & VAT$^{\ddagger\star}$ (ECCV'22)~\cite{hong2022cost} & 68.1 & 71.7 & 64.8 & \bf 63.3 & 67.0 & \bf 78.7 & \underline{72.6} & 74.1 & 69.5 & \bf 69.5 & \underline{71.4} & \underline{82.0}\\
        \cline{2-14}
         & SCCAN$^{\dagger}$ (Ours) & \underline{69.1} & \bf 74.0 & \underline{66.3} & 61.6 & \underline{67.7} & 77.3 & 71.6 & \underline{75.2} & 69.5 & 66.5 & 70.7 & 79.6\\
         & SCCAN$^{\ddagger}$ (Ours) & \bf 70.9 & \underline{73.9} & \bf 66.8 & 61.7 & \bf 68.3 & \underline{78.5} & \bf 73.1 & \bf 76.4 & \underline{70.3} & 66.1 & \bf 71.5 & \bf 82.1\\
        \hline
      \end{tabular}
  }
  \end{center}
  \caption{{\bf Comparison with state-of-the-arts on PASCAL-5$^i$}. {\bf Bold} results represent the best performance, while the \underline{underlined} results indicate the second best. $\star$ means a method originally uses different data lists for testing, and is adapted with the uniform ones. $\dagger$ and $\ddagger$ indicate that the resize methods from PFENet~\cite{tian2020prior} and HSNet~\cite{min2021hypercorrelation} are used (which is explained in supplementary material).}
  \label{tab:results_pascal}
\end{table*}

\begin{table*}[t]
  \begin{center}
  \resizebox{.9\linewidth}{!}{
      \begin{tabular}{c|r|cccccc|cccccc}
        \hline
        \multirow{2}{*}{Backbone} & \multirow{2}{*}{Method} & \multicolumn{6}{c|}{1-shot} & \multicolumn{6}{c}{5-shot}\\
        & & 20$^0$ & 20$^1$ & 20$^2$ & 20$^3$ & Mean & FB-IoU & 20$^0$ & 20$^1$ & 20$^2$ & 20$^3$ & Mean & FB-IoU \\
        \hline
        \multirow{9}{*}{ResNet50} & MLC$^{\dagger}$ (ICCV'21)~\cite{yang2021mining} & \bf 46.8 & 35.3 & 26.2 & 27.1 & 33.9 & - & \bf 54.1 & 41.2 & 34.1 & 33.1 & 40.6 & -\\
         & HSNet$^{\ddagger}$ (ICCV'21)~\cite{min2021hypercorrelation} & 36.7 & 41.4 & 39.5 & 39.1 & 39.2 & 67.6 & 44.4 & 49.7 & 46.1 & 45.5 & 46.4 & 70.9\\
         & CyCTR$^{\dagger}$ (NIPS'21)~\cite{zhang2021few} & 38.9 & 43.0 & 39.6 & 39.8 & 40.3 & - & 41.1 & 48.9 & 45.2 & 47.0 & 45.6 & -\\
         & NTRENet$^{\dagger}$ (CVPR'22)~\cite{liu2022learning} & 36.8 & 42.6 & 39.9 & 37.9 & 39.3 & 68.5 & 38.2 & 44.1 & 40.4 & 38.4 & 40.3 & 69.2\\
         & DPCN$^{\dagger}$ (CVPR'22)~\cite{liu2022dynamic} & - & - & - & - & 41.5 & 62.7 & - & - & - & - & - & -\\
         & DCAMA$^{\ddagger}$ (ECCV'22)~\cite{shi2022dense} & \underline{41.9} & 45.1 & 44.4 & 41.7 & 43.3 & \underline{69.5} & 45.9 & 50.5 & 50.7 & 46.0 & 48.3 & 71.7\\
         & VAT$^{\ddagger}$ (ECCV'22)~\cite{hong2022cost} & 39.1 & 43.5 & 42.1 & 39.9 & 41.1 & 68.2 & 45.2 & 50.1 & 48.0 & 45.6 & 47.2 & 71.6\\
        \cline{2-14}
         & SCCAN$^{\dagger}$ (Ours) & 39.5 & \underline{49.3} & \underline{47.3} & \underline{44.3} & \underline{45.1} & 68.5 & 45.7 & \underline{56.4} & \underline{56.5} & \underline{50.7} & \underline{52.3} & \underline{72.2}\\
         & SCCAN$^{\ddagger}$ (Ours) & 40.4 & \bf 49.7 & \bf 49.6 & \bf 45.6 & \bf 46.3 & \bf 69.9 & \underline{47.2} & \bf 57.2 & \bf 59.2 & \bf 52.1 & \bf 53.9 & \bf 74.2\\
        \hline
        \multirow{7}{*}{ResNet101} & PFENet$^{\dagger}$ (TPAMI'20)~\cite{tian2020prior} & 34.3 & 33.0 & 32.3 & 30.1 & 32.4 & 58.6 & 38.5 & 38.6 & 38.2 & 34.3 & 37.4 & 61.9\\
         & MLC$^{\dagger}$ (ICCV'21)~\cite{yang2021mining} & \bf 50.2 & 37.8 & 27.1 & 30.4 & 36.4 & - & \bf 57.0 & 46.2 & 37.3 & 37.2 & 44.4\\
         & HSNet$^{\ddagger}$ (ICCV'21)~\cite{min2021hypercorrelation} & 37.6 & 44.5 & 44.4 & 40.7 & 41.8 & 69.0 & 45.1 & 52.3 & 48.5 & 47.9 & 48.5 & 72.1\\
         & NTRENet$^{\dagger}$ (CVPR'22)~\cite{liu2022learning} & 38.3 & 40.4 & 39.5 & 38.1 & 39.1 & 67.5 & 42.3 & 44.4 & 44.2 & 41.7 & 43.2 & 69.6\\
         & DCAMA$^{\ddagger}$ (ECCV'22)~\cite{shi2022dense} & 41.5 & 46.2 & 45.2 & 41.3 & 43.5 & \bf 69.9 & 48.0 & 58.0 & 54.3 & 47.1 & 51.9 & 73.3\\
        \cline{2-14}
         & SCCAN$^{\dagger}$ (Ours) & 41.7 & \underline{51.3} & \underline{48.4} & \underline{46.7} & \underline{47.0} & 68.5 & 49.0 & \underline{59.3} & \underline{59.4} & \underline{52.7} & \underline{55.1} & \underline{73.4}\\
         & SCCAN$^{\ddagger}$ (Ours) & \underline{42.6} & \bf 51.4 & \bf 50.0 & \bf 48.8 & \bf 48.2 & \underline{69.7} & \underline{49.4} & \bf 61.7 & \bf 61.9 & \bf 55.0 & \bf 57.0 & \bf 74.8\\
        \hline
      \end{tabular}
  }
  \end{center}
  \caption{{\bf Comparison with state-of-the-arts on COCO-20$^i$}. {\bf Bold} results represent the best performance, while the \underline{underlined} results indicate the second best. $\dagger$ and $\ddagger$ indicate that the resize methods from PFENet~\cite{tian2020prior} and HSNet~\cite{min2021hypercorrelation} are used.}
  \label{tab:results_coco}
\end{table*}

\noindent
\textbf{Datasets.}
We evaluate our method on two public benchmark datasets, including PASCAL-5$^i$~\cite{shaban2017one} and COCO-20$^i$~\cite{nguyen2019feature}.
PASCAL-5$^i$ contains 20 classes, and is built upon PASCAL VOC 2012~\cite{everingham2010pascal} with additional annotations from SDS~\cite{hariharan2014simultaneous}, while COCO-20$^i$ is created from MSCOCO~\cite{lin2014microsoft}, which is more challenging and has 80 classes.
Both PASCAL-5$^i$ and COCO-20$^i$ are evenly split into 4 folds for cross validation, \ie, 5 and 20 classes per fold, respectively.
In each fold, the union of other three folds is for training, while the fold itself is preserved for testing.
Besides, 1,000 and 4,000 episodes are randomly sampled from PASCAL-5$^i$ and COCO-20$^i$ during testing.

\noindent
\textbf{Evaluation metrics.}
Following existing works~\cite{tian2020prior,zhang2021few,hong2022cost}, mean intersection over union (mIoU) and foreground-background IoU (FB-IoU) are adopted as evaluation metrics.
The former measures the mean IoU scores for all FG classes in a fold, while FB-IoU regards them as a single FG class, and reports the average IoU scores for FG and BG.

\noindent
\textbf{Implementation details.}
We take ResNet50/101~\cite{he2016deep} pretrained on ImageNet~\cite{russakovsky2015imagenet} as backbones, whose weights are frozen.
Following CyCTR~\cite{zhang2021few}, we use AdamW and SGD optimizers to optimize attention-related (\ie, SCCA) and other parameters, respectively, and Dice loss~\cite{milletari2016v} is taken as the loss function $\mathcal{L}_{Seg}$.
The model is trained for 200 epochs on PASCAL-5$^i$, and 50 epochs on COCO-20$^i$.
For both datasets, the batch size is fixed as 8, and the learning rates of AdamW and SGD are initialized as 6e-5 and 5e-3, respectively.
During training, all images are randomly cropped to 473$\times$473 patches as inputs, and we employ the same set of data augmentation techniques as PFENet~\cite{tian2020prior}.
During testing, we resize the outputs to compare with the original ground truths.
For SCCA, we employ 8 SCCA blocks, and set window size as 8. For other attention-related parameters, number of heads is 8, embedding dimension is 256, and MLP ratio is 1.
For $k$-shot setting, when $k>1$, we simply follow PFENet~\cite{tian2020prior} to average the support features.

\begin{figure*}[t]
  \begin{center}
  \includegraphics[width=1\linewidth]{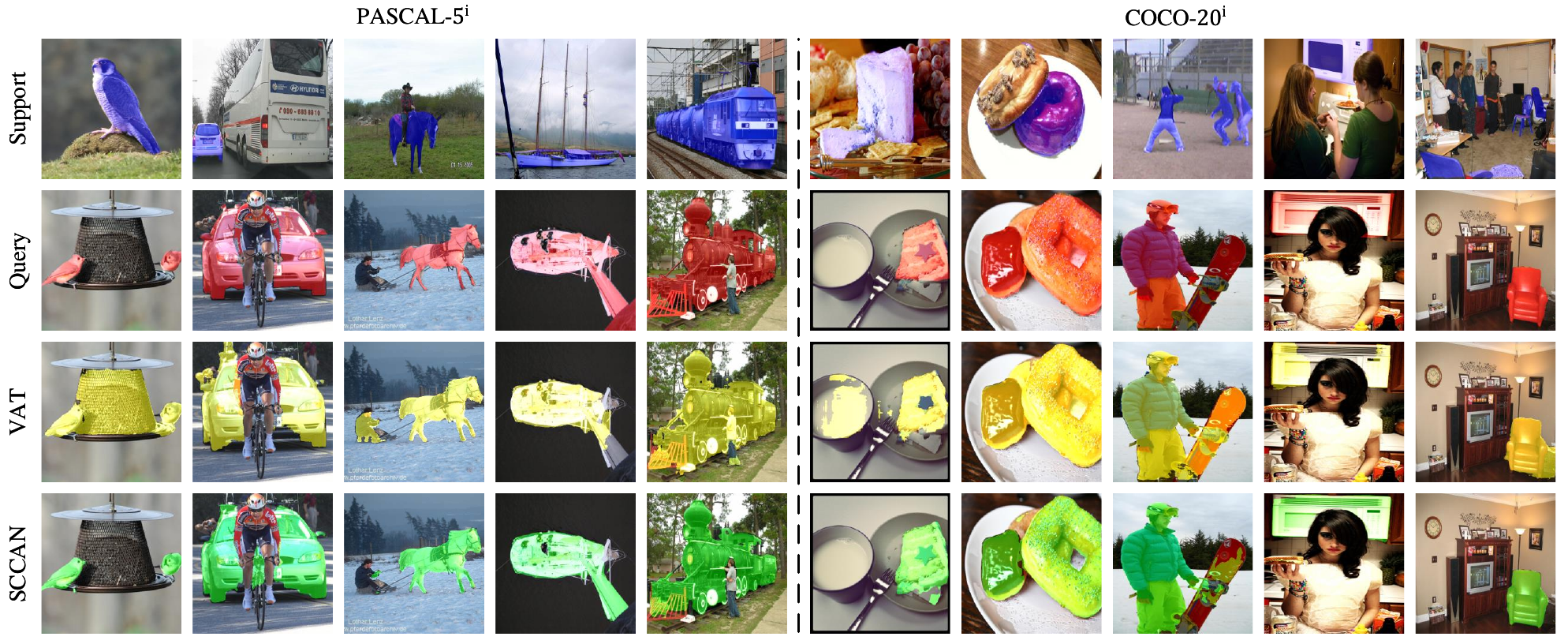}
  \end{center}
  \caption{{\bf Qualitative results of our SCCAN and VAT~\cite{hong2022cost} under 1-shot setting on PASCAL-5$^i$ and COCO-20$^i$}.}
  \label{fig:qualitative_results}
\end{figure*}

\subsection{Comparisons with state-of-the-arts}
\label{sec:comparisons_with_sotas}

\noindent
\textbf{Quantitative results.}
The results of baselines, as well as our proposed SCCAN, are shown in \cref{tab:results_pascal} and \cref{tab:results_coco}, and we could draw the following conclusions.
(1) SCCAN outperforms all baselines by considerable margins under most of the settings, and achieves new state-of-the-arts.
(2) Particularly, the gap between SCCAN and other methods is larger on COCO-20$^i$.
For example,
{\final with ResNet50, SCCAN outperforms others by 1.3\%+ and 3.0\%+ under 1-shot setting on two datasets, in terms of mIoU (averaged from 4 folds).}
When referring to 5-shot setting (with ResNet50), the gap is more prominent, \eg, mIoU of SCCAN is 5.6\%+ better than
{\final that of the best baseline DCAMA~\cite{shi2022dense}}
on COCO-20$^i$.
We believe this can be explained by the differences between two datasets.
PASCAL-5$^i$ contains many images with easy BG (\eg, pure sky), while COCO-20$^i$ is a more challenging benchmark, \eg, the images usually contain multiple FG objects from different classes.
In each episode, only one class is considered as FG, while other objects will be taken all as BG. Therefore, BG is much more complex in this case.
Our SCCAN directly mitigates the {\em BG mismatch} and {\em FG-BG entanglement} issues, and could deal with the complex BG better. Thus, SCCAN could gain more improvements on COCO-20$^i$ than on PASCAL-5$^i$.

\noindent
\textbf{Qualitative results.}
To better understand our SCCAN, we take some episodes from two datasets, then compare SCCAN with a latest baseline VAT~\cite{hong2022cost}, and the qualitative results are presented in \cref{fig:qualitative_results}.
We could observe that SCCAN is better at distinguishing FG and BG, \eg, in the third and fifth columns of PASCAL-5$^i$, VAT mistakenly classifies human as FG (horse and train), but SCCAN could recognize them well. Moreover, SCCAN could sometimes yield more reasonable results than human-annotated labels, \eg, in the second column of COCO-20$^i$, there exist a hole for plate in the right doughnut, SCCAN could point it out, but VAT and ground truth wrongly consider it as FG.

\subsection{Ablation study}
\label{sec:ablation_study}

\begin{table}[t]
  \begin{center}
  \scalebox{0.85}{
      \begin{tabular}{cccccc}
        \hline
        PMA & Attention & PA & SC & mIoU & FB-IoU\\
        \hline
         &  &   &  & 62.3 & 73.3\\
        \checkmark &   &  &  & 65.2 & 76.0\\
        \checkmark & ST  &  &  & 65.9 & 75.5\\
        \checkmark & CST  &  &  & 65.3 & 75.4\\
        \checkmark & SCCA &  &  & 66.0 & 75.3\\
        \checkmark & SCCA & \checkmark &  & \underline{66.5} & \underline{76.2}\\
        \checkmark & SCCA & \checkmark & \checkmark & \bf 67.0 & \bf 76.4\\
        \hline
      \end{tabular}
  }
  \end{center}
  \caption{{\bf Ablation study}. mIoU shows the average score of 4 folds. {\bf PMA}: Pseudo mask aggregation; {\bf ST}: Swin transformer; {\bf CST}: Swin transformer with cross attentions; {\bf SCCA}: Self-calibrated cross attention; {\bf PA}: Patch alignment; {\bf SC}: Scaled-cosine.}
  \label{tab:ablation_components}
  \vspace{-1em}
\end{table}

\noindent
\textbf{Components analysis.}
To validate the effectiveness of proposed modules, we perform ablation study with ResNet50 on PASCAL-5$^i$, under 1-shot setting.
SCCAN mainly consists of PMA module and SCCA which further contains PA, and SC (Scaled-cosine) mechanism.
Our basic model is obtained when we remove all these components from SCCAN.
As shown in \cref{tab:ablation_components}, the usage of PMA could bring 2.9\% growth of mIoU, and if we further employ the standard ST for feature enhancement, the mIoU score could reach 65.9\%.
When we merely replace self attention as cross attention in ST (denoted as CST), the {\em BG mismatch} and {\em FG-BG entanglement} issues would arise, and the mIoU score will be 0.6\% worse.
Fortunately, SCCA could mitigate these issues, but the obtained score is close to that of the standard ST (66.0\% vs 65.9\%), which is caused by {\em invalid\&misaligned support patch} issues (described in \cref{sec:self_calibrated_cross_attention}). PA is used to tackle the issues, and the score is 66.5\%.
Finally, SC mechanism aims at encouraging cross attention, and the final mIoU is boosted to 67.0\%.

\begin{figure}[t]
  \begin{center}
  \includegraphics[width=1\linewidth]{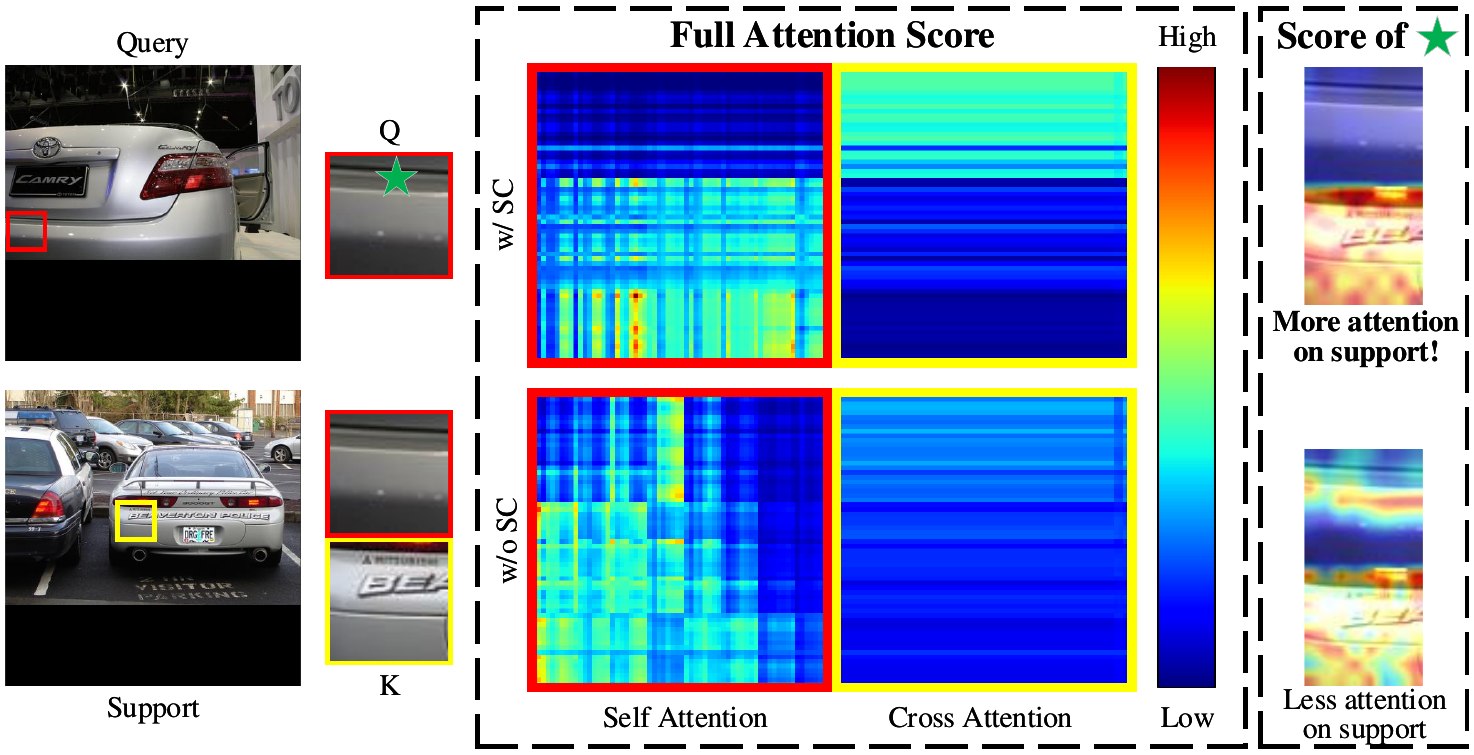}
  \end{center}
  \caption{{\bf Visualization of attention scores (after softmax) with/without SC mechanism}. Zoom in for more details. Red rectangle means query patch, while yellow rectangle means support patch.}
  \label{fig:sc_visualization}
\end{figure}

\noindent
\textbf{Visualization of scaled-cosine mechanism.}
To better understand the function of SC mechanism, we refer to the last two rows of \cref{tab:ablation_components}, and visualize their attention scores with an example in \cref{fig:sc_visualization}.
The yellow support patch is the aligned patch of the red one.
{\xu
Referring to the rightmost column in ~\cref{fig:sc_visualization}, we could observe that for a query FG pixel, SC mechanism could help to focus more on support FG features, \ie, more information could be obtained from support patch.
}
In this way, support information is better utilized.

\begin{table}[t]
  \begin{center}
  \scalebox{0.85}{
      \begin{tabular}{ccccccc}
        \hline
        Patch size & 5$^0$ & 5$^1$ & 5$^2$ & 5$^3$ & mIoU & FB-IoU\\
        \hline
        5  & 67.1 & 72.4 & 66.5 & 60.0 & 66.5 & 76.0\\
        8  & 67.5 & \bf 72.6 & \bf 67.2 & \bf 60.5 & \bf 67.0 & \bf 76.4\\
        10 & 67.2 & 72.3 & 66.8 & 60.1 & 66.6 & 75.6\\
        15 & \bf 67.8 & 72.5 & 66.4 & 59.8 & 66.6 & 75.1\\
        \hline
      \end{tabular}
  }
  \end{center}
  \caption{{\bf Parameter study of window/patch sizes on PASCAL-5$^i$}. mIoU shows the average score of 4 folds.}
  \label{tab:window_sizes}
  \vspace{-1em}
\end{table}

\noindent
\textbf{Patch/Window size in SCCA.}
Patch size plays an important role in swin transformer. With the decrease of patch size, less GPU memory is required, but one pixel could access to less pixels within one attention.
In SCCA, patch size would also affect patch alignment. Therefore, we take values from \{5, 8, 10, 15\} to study its impacts.
As shown in \cref{tab:window_sizes}, when patch size is 8, best performance is achieved.

\noindent
\textbf{Number of SCCA blocks.}
SCCA blocks are the basis of our proposed SCCAN, and we conduct experiments to study SCCAN with different number of SCCA blocks. As shown in ~\cref{tab:num_scca}, the best mIoU and FB-IoU scores could be achieved when we use 8 SCCA blocks. The Floating Point Operations (FLOPs) score is calculated based on one query and support image pairs, whose shapes are uniformly set as $473\times473$.

\begin{table}[t]
  \begin{center}
  \scalebox{0.85}{
      \begin{tabular}{ccccc}
        \hline
        \#SCCA Blocks & mIoU & FB-IoU & \#Params (M) & FLOPs (G) \\
        \hline
        4  & 66.3 & 75.8 & 31.8 & 447.7 \\
        8  & \bf 67.0 & \bf 76.4 & 35.0 & 480.9 \\
        12 & 66.3 & 76.0 & 38.1 & 514.1 \\
        16 & 66.2 & 75.8 & 41.3 & 547.3 \\
        \hline
      \end{tabular}
  }
  \end{center}
  \caption{{\bf Parameter study of number of SCCA blocks on PASCAL-5$^i$}. mIoU shows the average score of 4 folds.}
  \label{tab:num_scca}
  \vspace{-1em}
\end{table}

\section{Conclusion}
\label{sec:conclusion}

We propose a self-calibrated cross attention network (SCCAN) for accurate FSS, which consists of SCCA blocks and PMA module.
SCCA is designed to effectively fuse query features with support FG features.
PMA could generate robust pseudo masks for query images with minor cost.
Extensive experiments are conducted and show that SCCAN could achieve new state-of-the-arts for FSS.

\smallskip
\noindent
\textbf{Limitation.}
Currently, we follow PFENet~\cite{tian2020prior} to average the features of 5 support samples under $5$-shot setting. However, cross attention will be affected, \eg, if there are objects at the top-left and bottom-right corners for 1 and 4 support images, respectively, then their features will roughly multiply 1/5 and 4/5 due to the average operation, and the top-left support features will be less similar to query features, and less used. Thus, a $k$-shot strategy should be particularly designed for cross attention, and we leave it for future work.

\smallskip
\noindent
\textbf{Acknowledgement.}
This study is supported under the RIE2020 Industry Alignment Fund – Industry Collaboration Projects (IAF-ICP) Funding Initiative, as well as cash and in-kind contribution from the industry partner(s).

{\small
\bibliographystyle{ieee_fullname}
\bibliography{egbib}
}

\newpage
\appendix



{\xu
\section{Other experiments}
\label{sec:other_experiments}

\subsection{Difficult query samples}
\label{sec:difficult_query_samples}

\begin{figure}[h]
  \begin{center}
  \includegraphics[width=\linewidth]{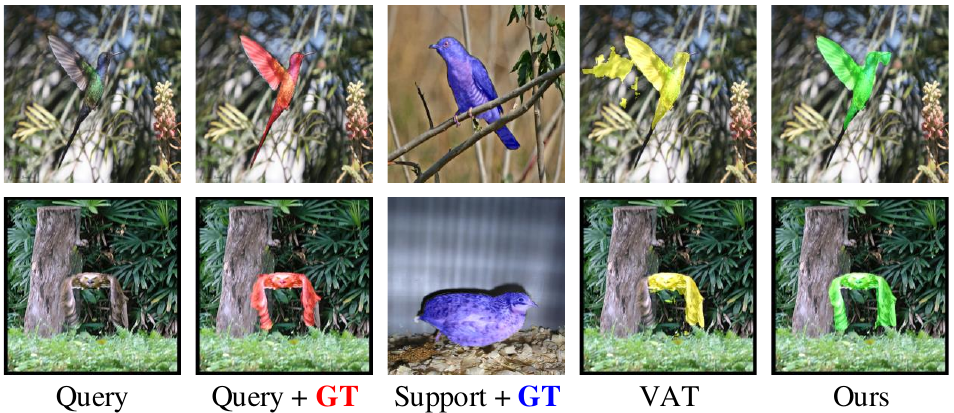}
  \end{center}
  \caption{{\bf Difficult query samples where query FG and query BG look similar.} Zoom in for more details.}
  \label{fig:difficult_query_samples}
\end{figure}

As shown in \cref{fig:difficult_query_samples}, we further provide two difficult examples. We could observe that query FG objects look very similar to query BG, and it is even hard for human to perform segmentation. VAT~\cite{hong2022cost} could not accurately separate query FG and BG in this case, while our model could mitigate the {\em FG-BG entanglement} issue well, and the segmentation results are quite good.




\subsection{Weak support annotations}
\label{sec:weak_support_annotations}

To further reduce annotation costs of support images, we follow existing studies~\cite{wang2019panet,zhang2019canet,liu2022dynamic} to use cheaper bounding box annotations. As shown in \cref{tab:weak_support_annotations}, SCCAN could outperform others well, and the performance is slightly worse than using expensive pixel-wise masks, which validates the effectiveness of SCCAN, \ie, query FG and BG could be well differentiated.

\begin{table}[h]
  \begin{center}
  \scalebox{0.85}{
      \begin{tabular}{rcccccc}
        \hline
        Method & 5$^0$ & 5$^1$ & 5$^2$ & 5$^3$ & Mean & FB-IoU\\
        \hline
        PANet$^{\dagger}$~\cite{wang2019panet} & - & - & - & - & 45.1 & -\\
        CANet$^{\dagger}$~\cite{zhang2019canet} & - & - & - & - & 52.0 & -\\
        DPCN$^{\dagger}$~\cite{liu2022dynamic} & 59.8 & 70.5 & 63.2 & 55.5 & 62.3 & -\\
        SCCAN$^{\dagger}$ & \underline{67.3} & \underline{71.8} & \underline{65.6} & \underline{58.0} & \underline{65.7} & \underline{75.5}\\
        SCCAN$^{\ddagger}$ & \bf 67.5 & \bf 72.6 & \bf 67.2 & \bf 60.5 & \bf 67.0 & \bf 76.4\\
        \hline
      \end{tabular}
  }
  \end{center}
  \caption{{\bf Study on weak support annotations}. Support annotations are whittled from pixel-wise masks to bounding boxes. DPCN utilizes multi-scale testing. $\dagger$ means using bounding boxes, $\ddagger$ means using pixel-wise masks.}
  \label{tab:weak_support_annotations}
\end{table}

\subsection{GPU memory cost}
\label{sec:gpu_memory_cost}

We further show the GPU memory cost of our SCCAN (with 8 SCCA blocks), CyCTR~\cite{zhang2021few} and VAT~\cite{hong2022cost} in \cref{tab:memory_cost}. Compared with other attention-based methods, we could observe that our SCCAN could save much GPU memory, which could demonstrate the effectiveness of our design.

\begin{table}[h]
  \begin{center}
  \scalebox{0.85}{
      \begin{tabular}{cccc}
        \hline
        Method & Input shape & GPU memory (Mb) \\
        \hline
        CyCTR (NIPS'21)~\cite{zhang2021few} & \multirow{3}{*}{$4\times3\times473\times473$} & 25,463 \\
        VAT (ECCV'22)~\cite{hong2022cost} & & 23,553 \\
        SCCAN (Ours) & & \bf 10,667 \\
        \hline
      \end{tabular}
  }
  \end{center}
  \caption{{\bf GPU memory cost of different methods}. We uniformly set the batch size as 4, and set the image size as $473\times473$.}
  \label{tab:memory_cost}
  \vspace{-1em}
\end{table}

\subsection{More testing episodes on COCO-20$^i$}
\label{sec:more_test_coco}

\begin{table*}[t]
  \begin{center}
  \resizebox{.9\linewidth}{!}{
      \begin{tabular}{c|r|cccccc|cccccc}
        \hline
        \multirow{2}{*}{Backbone} & \multirow{2}{*}{Method} & \multicolumn{6}{c|}{1-shot} & \multicolumn{6}{c}{5-shot}\\
        & & 20$^0$ & 20$^1$ & 20$^2$ & 20$^3$ & Mean & FB-IoU & 20$^0$ & 20$^1$ & 20$^2$ & 20$^3$ & Mean & FB-IoU \\
        \hline
        \multirow{5}{*}{ResNet50} & HSNet$^{\ddagger}$ (ICCV'21)~\cite{min2021hypercorrelation} & 35.1 & 41.8 & 38.8 & 38.4 & 38.5 & 65.4 & 41.9 & 49.8 & 46.7 & 44.3 & 45.7 & 69.0\\
         & DCAMA$^{\ddagger}$ (ECCV'22)~\cite{shi2022dense} & 38.3 & 41.9 & 43.3 & 40.0 & 40.9 & 63.5 & 43.5 & 49.2 & 49.6 & 46.3 & 47.2 & 66.2\\
         & VAT$^{\ddagger}$ (ECCV'22)~\cite{hong2022cost} & 36.5 & 43.2 & 41.3 & 38.9 & 40.0 & 66.2 & 41.9 & 49.7 & 48.3 & 44.4 & 46.1 & 69.4\\
        \cline{2-14}
         & SCCAN$^{\dagger}$ (Ours) & \underline{38.5} & \underline{49.1} & \underline{45.4} & \underline{43.7} & \underline{44.2} & \underline{68.1} & \underline{45.4} & \underline{54.7} & \underline{52.7} & \underline{50.7} & \underline{50.9} & \underline{71.8}\\
         & SCCAN$^{\ddagger}$ (Ours) & \bf 39.8 & \bf 50.2 & \bf 47.7 & \bf 45.7 & \bf 45.8 & \bf 69.7 & \bf 47.6 & \bf 57.7 & \bf 57.5 & \bf 53.0 & \bf 59.9 & \bf 74.2\\
        \hline
        \multirow{4}{*}{ResNet101} & HSNet$^{\ddagger}$ (ICCV'21)~\cite{min2021hypercorrelation} & 35.9 & 44.8 & 41.4 & 40.9 & 40.8 & 66.0 & 43.2 & 51.4 & 48.9 & 47.3 & 47.7 & 69.9\\
         & DCAMA$^{\ddagger}$ (ECCV'22)~\cite{shi2022dense} & 39.7 & 46.0 & 45.2 & 40.2 & 42.8 & 64.2 & 45.7 & 54.7 & 52.7 & 46.4 & 49.9 & 67.5\\
        \cline{2-14}
         & SCCAN$^{\dagger}$ (Ours) & \underline{41.0} & \underline{51.5} & \underline{46.9} & \underline{46.1} & \underline{46.4} & \underline{68.3} & \underline{47.8} & \underline{58.5} & \underline{56.6} & \underline{53.4} & \underline{54.1} & \underline{73.2}\\
         & SCCAN$^{\ddagger}$ (Ours) & \bf 42.3 & \bf 52.2 & \bf 49.5 & \bf 47.9 & \bf 48.0 & \bf 69.8 & \bf 49.4 & \bf 61.4 & \bf 60.2 & \bf 55.8 & \bf 56.7 & \bf 74.8\\
        \hline
      \end{tabular}
  }
  \end{center}
  \caption{{\bf Testing results of 20,000 episodes on COCO-20$^i$}. {\bf Bold} results represent the best performance, while the \underline{underlined} results indicate the second best. $\dagger$ and $\ddagger$ indicate that the resize methods from PFENet~\cite{tian2020prior} and HSNet~\cite{min2021hypercorrelation} are used during testing, respectively.}
  \label{tab:more_test_coco}
\end{table*}

To make test results more reliable, we follow PFENet~\cite{tian2020prior} to randomly sample 20,000 episodes from COCO-20$^i$ to perform meta-test again, and show the results in \cref{tab:more_test_coco}.
As we could observe from the table, our SCCAN could still outperform previous state-of-the-arts by large margins, including HSNet~\cite{min2021hypercorrelation}, DCAMA~\cite{shi2022dense} and VAT~\cite{hong2022cost}.
}

\section{Discussions}
\label{sec:discussions}

\subsection{Support background utilization}
\label{sec:support_bg_utilization}

Recall that we target on cross attention methods for FSS in this paper, \ie, we aim to use cross attention to combine query features with support FG features.
However, existing methods suffer from two issues, namely, {\em BG mismatch} and {\em FG-BG entanglement}, both of which are raised due to the fact that query BG cannot find matched features in support FG. As a result, query BG will inevitably be fused with dissimilar features and get biased. In addition, query FG also correctly aggregate matched support FG features, and as both query FG and BG take in support FG, they get entangled, which is against the goal of FSS, \ie, distinguish query FG and BG.

Naturally, a na\"ive idea is to find matched BG features from support BG. Nevertheless, we claim that it is not appropriate to use support BG in cross attention-based FSS, which is explained as follows.

\begin{figure}[h]
  \begin{center}
  \includegraphics[width=0.7\linewidth]{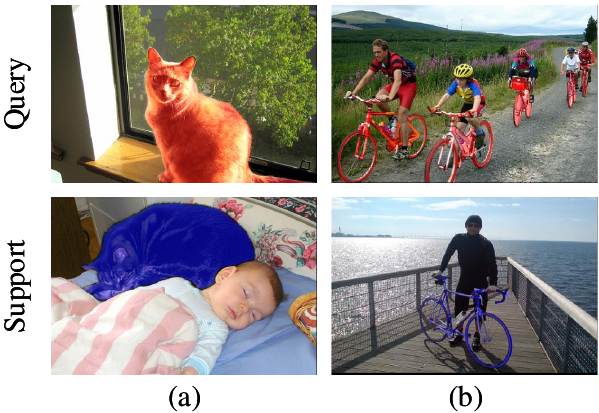}
  \end{center}
  \caption{{\bf Dissimilar background of query and support images}.}
  \label{fig:att_issues}
\end{figure}

(1) Support BG is not always similar to query BG, \eg, in \cref{fig:att_issues}(a), two cats in query and support images are surrounded by window and bed, respectively; while in \cref{fig:att_issues}(b), the background objects for bicycle are mountain and sea, respectively. As query BG still cannot find matched features in support samples, the {\em BG mismatch} and {\em FG-BG entanglement} issues remain.

\begin{figure}[h]
  \begin{center}
  \includegraphics[width=\linewidth]{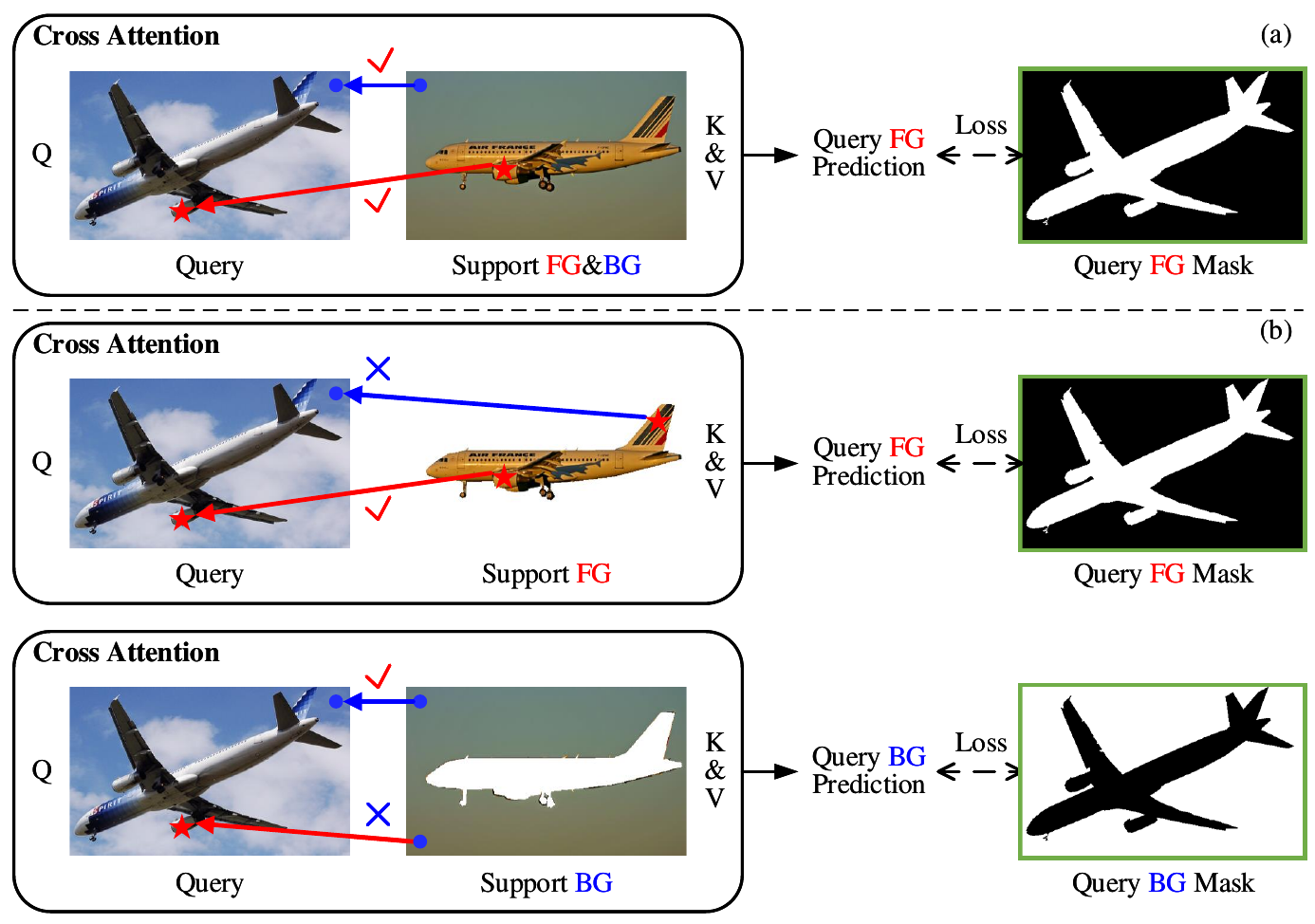}
  \end{center}
  \caption{{\bf Two possible ways of using support BG. (a) Single cross attention. (b) Double cross attentions.}}
  \label{fig:cross_attention_support_bg}
  \vspace{-1em}
\end{figure}

\begin{figure*}[t]
  \begin{center}
  \includegraphics[width=\linewidth]{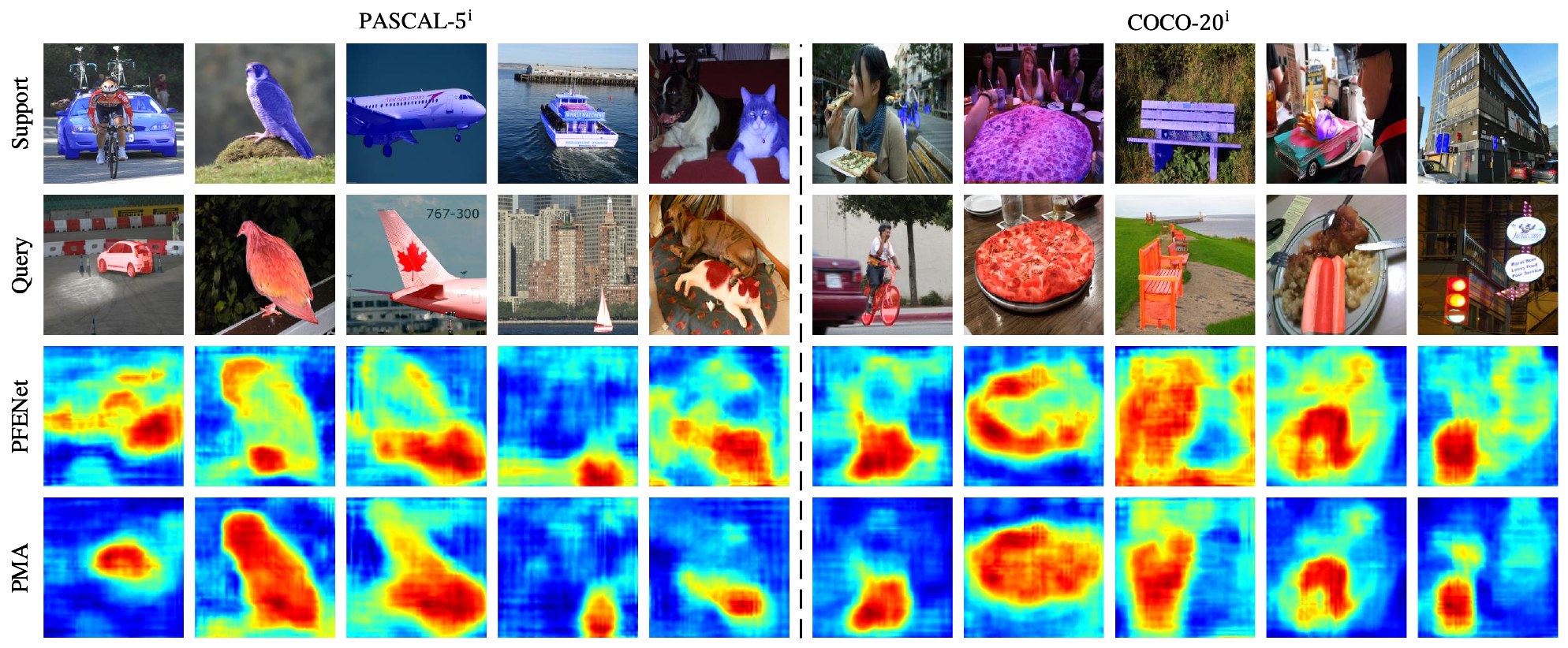}
  \end{center}
  \caption{{\bf Comparisons of training-agnostic pseudo mask generation methods between PFENet~\cite{tian2020prior} and our proposed pseudo mask aggregation (PMA) module.}
  }
  \label{fig:pseudo_mask_aggregation_pfenet}
  \vspace{-1em}
\end{figure*}

(2) Even though support and query BG could be similar, cross attention will still fail.
There are roughly two attempts:
(a) Use the complete support features as $K\&V$. Although cross attention works normally now (as in \cref{fig:cross_attention_support_bg}(a)), this way is against the expectation of FSS, \ie, the model could not know which part in the support sample is the region of interest (ROI). Specifically, the sky in query is similar to that in support, and the planes in two images are also similar. Recall that in FSS, different objects in query image will be considered as FG, with the changing of support information, \eg, if only sky is provided in support, then the model is expected to extract the sky-related pixels from query and consider them as FG. Thus, we cannot use the whole support features as $K\&V$.
(b) Jointly train two cross attentions. As shown in \cref{fig:cross_attention_support_bg}(b), two rows aim at extracting plane and sky, respectively. Although the problem in (a) is solved, the aforementioned {\em BG mismatch} and {\em FG-BG entanglement} come back, \eg, in two rows, query BG could not find matched BG in support FG, and query FG cannot find its matched FG in support BG, respectively.

In summary, support BG cannot be effectively used in cross attention-based FSS to mitigate our proposed issues, and our solution is absolutely effective and novel.

\subsection{Pseudo mask aggregation and PFENet}
\label{sec:pseudo_mask_aggregation_pfenet}

Pseudo/Prior masks are recently popular in FSS, which could roughly locate query FG objects without learnable parameters, by measuring similarity between high-level query and support features (with annotations) that are directly obtained from the pretrained backbone.

PFENet~\cite{tian2020prior} firstly propose such cheap but effective mechanism. Specifically, it firstly measures the similarity between each pair of query pixels and support FG pixels. Then, for each query pixel, its largest similarity score is normalized and taken as its probability of being query FG.

However, there exist two issues:
(1) PFENet only uses support FG for comparison, which will be limited when the gap between query FG and support FG is large, \eg, human head (query FG), and human arms (support FG). Consequently, the similarity scores between support FG and query FG/BG are both small, and thus, the locating function of pseudo masks are weakened. For instance, in the forth column of \cref{fig:pseudo_mask_aggregation_pfenet}, the ships in query and support are not so similar. Although the generated mask could still locate the ship in query, there are many wrongly activated BG pixels, and the model is likely to also take them as FG.
(2) As PFENet only takes each query pixel's largest similarity score for reference, the generated pseudo mask is not robust, \eg, it will be heavily affected by noises. As we could observe from the first column of \cref{fig:pseudo_mask_aggregation_pfenet}, the pseudo mask of PFENet cannot locate the car in query image well.

As introduced in the paper, PMA could mitigate these issues by:
(1) We also take support BG into consideration, \eg, human head (query FG) is more similar to human arms (support FG), compared with room (support BG).
(2) For each query pixel, we calculate its similarity scores with all support pixels, including FG and BG. Then, the normalized scores are used to aggregate the support mask values (FG is 1, BG is 0) and generate the pseudo mask. In this way, the side-effect of single largest value could be suppressed.

{\final
Some training-agnostic pseudo masks obtained from PFENet~\cite{tian2020prior} and our PMA module are visualized in \cref{fig:pseudo_mask_aggregation_pfenet}.
Besides, we further use threshold 0.75 to binarize the pseudo masks and directly measure their mIoU scores (averaged from 4 folds) on PASCAL-5$^i$, with ResNet50 as the backbone. The scores of pseudo masks from PFENet and PMA are 22.8\% and 38.7\%, respectively.
In conclusion,
}
PMA could consistently outperform PFENet in two aspects:
(1) PMA is better at locating query FG.
(2) There are less wrongly activated BG pixels in the generated pseudo masks.

{\xu
\subsection{Comparison with CyCTR~\cite{zhang2021few}}
\label{sec:compare_cyctr}

{\final
It is confusing that CyCTR~\cite{zhang2021few} seems to be a solution to the {\em BG mismatch} issue, but it is actually not, and we explain the reasons as follows.
(1) It is more appropriate to claim that CyCTR is likely to not have the above issue, when {\em query and support {\bf BG}} belong to the {\em same class}. However, we mention the side-effects of using support BG in cross attention-like methods in ~\cref{sec:support_bg_utilization}.
(2) When {\em {\bf BG} classes differ}, CyCTR consistently suffers from the above issue.
Its cycle-consistent attention does not solve this issue, which starts from a support BG pixel $P_S$, finds its most similar query pixel $P_Q$, and find $P_Q$'s most similar support pixel $P_S^{\prime}$. $P_S$ is preserved as long as $P_S^{\prime}$ belongs to BG, regardless of the difference between {\bf BG} classes, \ie, its cross attention still fuses {\em dissimilar} support BG to query BG.
}

\subsection{Comparison with VAT~\cite{hong2022cost}}
\label{sec:compare_vat}

VAT~\cite{hong2022cost} looks similar to our SCCAN in the following two aspects:
(1) It is also built upon swin transformer~\cite{liu2021swin}.
(2) VAT converts FSS task to semantic correspondence task which focuses on features matching and fusion.

However, VAT and existing cross attention FSS methods~\cite{wang2020few,zhang2019pyramid,zhang2021few} only target at the matching of {\bf query FG}, and suffer from the {\em BG mismatch} and {\em FG-BG entanglement} issues. Instead, our main purpose is to match and fuse {\bf query BG} with appropriate BG features to mitigate these issues.

\subsection{Comparison with BAM~\cite{lang2022learning}}
\label{sec:compare_bam}

BAM~\cite{lang2022learning} is a latest baseline, however, we do not include it in the main results because BAM adopts a special setting, which is a bit different from the standard one. That is, BAM extends standard FSS methods by using base classes' segmentation results, and its meta learner could be any standard FSS model (including ours).

In spite of the setting difference, with ResNet-50, our model (70.3\%) could still achieve comparable 5-shot results with BAM (70.9\%) on PASCAL-5$^i$, and our model (52.3\%) surpasses BAM (51.2\%) on COCO-20$^i$.

\subsection{Comparison with SSP~\cite{fan2022self}}
\label{sec:compare_ssp}

SSP~\cite{fan2022self} also focuses on the matching issues, and we explain the differences, as well as the superiority of our model, as follows:

\begin{figure}[h]
  \begin{center}
  \includegraphics[width=\linewidth]{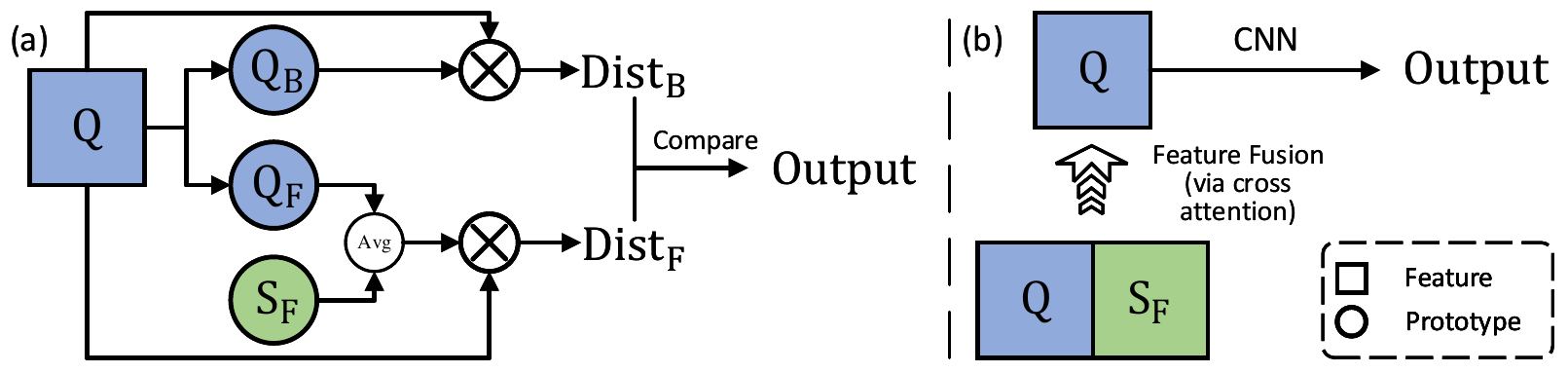}
  \end{center}
  \caption{{\bf Frameworks (a) SSP and (b) SCCAN.}}
  \label{fig:compare_ssp}
\end{figure}

(1) As illustrated in \cref{fig:compare_ssp}(a), SSP coarsely separates query FG and BG first, based on which FG and BG prototypes are then constructed. Finally, similarities are calculated between query features and two prototypes for segmentation. Instead, our model (as shown in \cref{fig:compare_ssp}(b)) enhances query features with the concatenation of query features and support FG, and then use CNN for segmentation.

(2) SSP's BG prototype is unreliable.
(a) With the coarse separation in the first phase, we cannot guarantee if the obtained BG prototype contains pure BG information or not;
(b) Query BG usually contains multiple objects, \eg, tree and grass in the second row of \cref{fig:difficult_query_samples}. As BG prototype only contains features that are most dissimilar to support FG, it may only include grass features in this case. Hence, tree features and BG prototype still mismatch.

(3) Our model outperforms SSP by a large margin, \eg, 1-shot mIoU on PASCAL-5$^i$ is 6.1\% better than that of SSP.
}

\section{Different resize methods}
\label{sec:resize_methods}

\begin{figure}[t]
  \begin{center}
  \includegraphics[width=\linewidth]{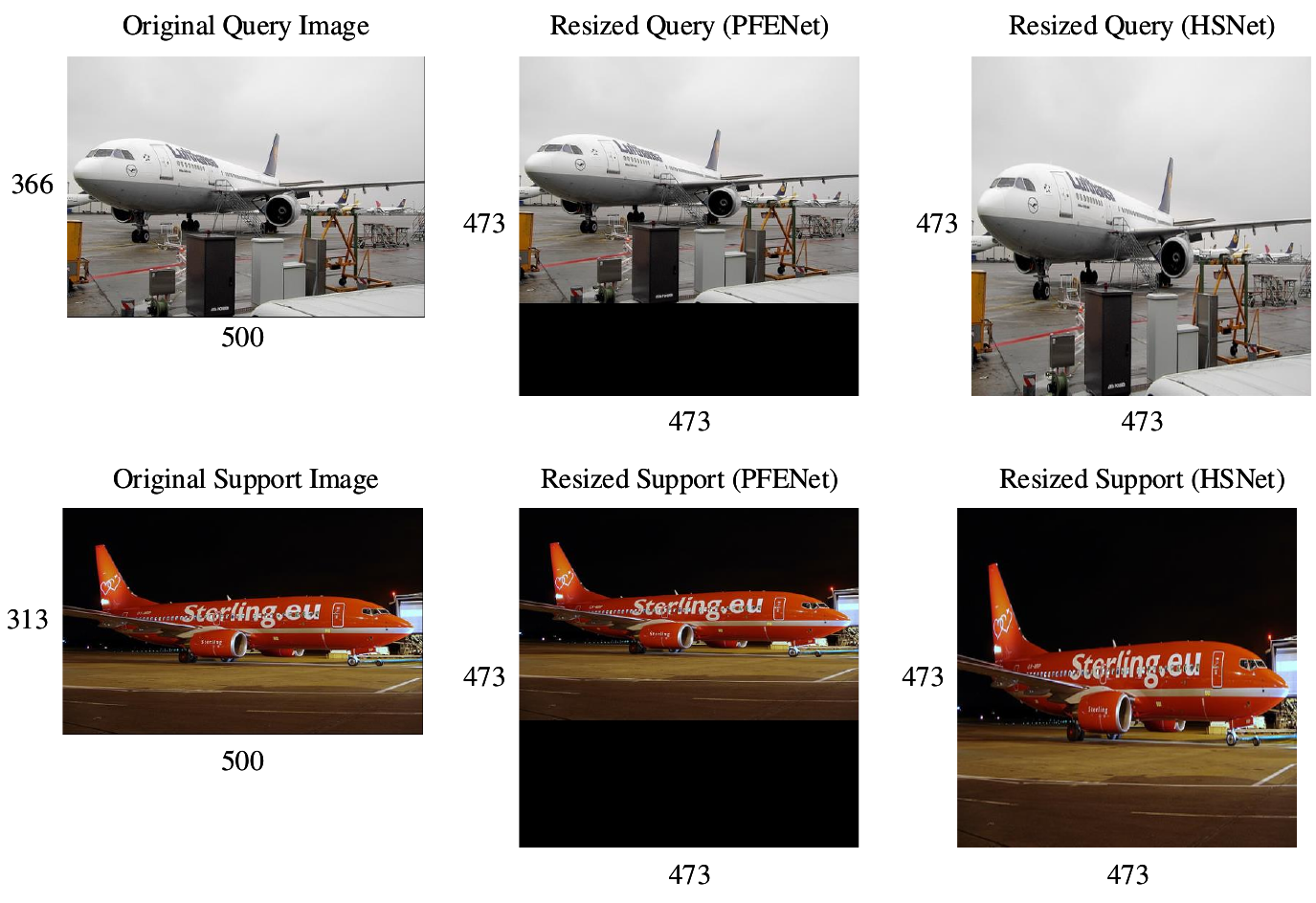}
  \end{center}
  \caption{{\bf Different resize methods adopted in existing baselines. (a) Original aspect ratio. (b) Different aspect ratio.}}
  \label{fig:resize_methods}
\end{figure}

As mentioned in the tables of quantitative results, there are two resize methods in existing state-of-the-arts methods, and we illustrate their difference in \cref{fig:resize_methods}.
In short summary, the resize method used in PFENet~\cite{tian2020prior} will resize the input query and input images, while keeping the original aspect ratio, and the shorter edge will be padded. Instead, HSNet~\cite{min2021hypercorrelation} directly resizes the images to the specific shape, \eg, 473$\times$473.
As a result, the resized objects in HSNet will be larger than those in PFENet. Particularly, HSNet could access to better support information, and the obtained segmentation results could be better.

\end{document}